\documentclass{article}
\usepackage{spconf,amsmath,graphicx}
 \usepackage{xcolor}
  
   \usepackage{graphicx}

\usepackage{makecell}

\usepackage{enumitem}
\usepackage{amsmath,amsfonts}
\setlist[enumerate]{itemsep=0pt,parsep=0pt,topsep=0pt}
\usepackage{hyperref}
\usepackage{float}

\title{Spectral-Spatial Contrastive Learning Framework for Regression on Hyperspectral Data}
\name{Mohamad Dhaini\textsuperscript{1,2 *}, Paul Honeine\textsuperscript{1}, Maxime Berar\textsuperscript{1}, Antonin Van Exem\textsuperscript{2} \thanks{* Corresponding Author \newline This work is funded by Tellux Company and LITIS UR 4108 lab.}}

\address{\textsuperscript{1} 
Univ Rouen Normandie, INSA Rouen Normandie, Université Le Havre Normandie, \\
Normandie Univ, LITIS UR 4108, F-76000 Rouen, France
\\
\textsuperscript{2} Tellux, 76150 Maromme, France}

\begin{document}
%
\maketitle
\begin{abstract}
Contrastive learning has demonstrated great success in representation learning, especially for image classification tasks. However, there is still a shortage in studies targeting regression tasks, and more specifically applications on hyperspectral data. In this paper, we propose a spectral-spatial contrastive learning framework for regression tasks for hyperspectral data, in a model-agnostic design allowing to enhance backbones such as 3D convolutional and transformer-based networks. Moreover, we provide a collection of transformations relevant for augmenting hyperspectral data. Experiments on synthetic and real datasets show that the proposed framework and transformations significantly improve the performance of all studied backbone models. 

\end{abstract}
\begin{keywords}
Contrastive Learning, Hyperspectral Data, Regression, Spectral-spatial Data Augmentation
\end{keywords}
\section{Introduction}
\label{sec:intro}
Hyperspectral imagery offers valuable insights into the physical properties of an object or area without the need for physical contact. Hyperspectral sensors capture a broad range of information within the light spectrum, often spanning hundreds of contiguous bands across a wavelength range of approximately 500 nm to 2500 nm. This capability allows each material to possess its own distinct spectral signature. This type of data has been gaining significant attention from the signal processing and machine learning community, mainly for 
classification \cite{roy2019hybridsn} 
and unmixing \cite{dhaini2022end} tasks.

Recently, due to the limitation of supervised learning techniques to labeled data,
self-supervised learning methods have been gaining  popularity to learn general representations from unlabeled data. 
In this context, contrastive learning is a self-supervised approach that provides such a relevant representation. For this purpose, the latent space representation is learned through a contrastive loss, by maximizing the agreement between differently augmented views of the same sample \cite{chen2020simple,braham2025spectralearth}.


In our related work \cite{dhaini2024contrastive}, spectral contrastive learning for hyperspectral data has demonstrated its efficiency, providing robustness to the neural network to several kinds of spectral variabilities and yielding robust and reliable results across real and synthetic applications. However, in real-world scenarios, hyperspectral data is not limited to just spectral variabilities; it also contains critical spatial information. The spatial aspect of the data represents the structure and relationship of features within a given space, offering a rich context that spectral data alone might overlook \cite{shenming2022new}. Incorporating spatial information could lead to significant advancements, such as better generalization scores by leveraging the spatial relationships and patterns within the data, enhanced noise reduction capabilities through the identification of spatial outliers, and a deeper understanding of the contextual interactions between different features.
This paves the way to the idea of investigating the potential of spatial contrastive learning.  

In this paper, we capitalize on the strengths of both spectrally and spatially methods, by exploring the fusion of spectral-spatial contrastive learning. We also consider the more difficult regression tasks. Therefore, the main contributions can be seen as following:
\begin{enumerate}
    \item We revisit some spatial augmentation techniques, often used in computer vision, to fit into hyperspectral data. Moreover, we make use of the spectral transformations to create a spectral-spatial toolbox.
    \item We propose a framework suitable for spectral-spatial contrastive learning on hyperspectral data with the use of 3D-based backbone architectures, including 3D CNN and transformer-based architectures.
    \item We show the relevance of our framework in enhancing 5 recent hyperspectral methods in synthetic and real-world scenarios, including unmixing and prediction of pollution concentration in soil data.
\end{enumerate}
The rest of the paper is organized as follows. Section \ref{sec:related} highlights some of the related work on contrastive learning with hyperspectral data. In Section \ref{sec:method}, we present the core ideas behind our proposed method. The experimental studies with the obtained results are presented in Section \ref{sec:results}. The contributions and future steps are summarized in Section \ref{sec:conclusion}.

\section{Related Work}
\label{sec:related}

Recently, there have been some studies investigating the use of contrastive learning on hyperspectral data. However, the majority of these studies were targeting classification tasks. In \cite{wang2023nearest} a contrastive learning network is introduced based on a nearest neighbor augmentation scheme, 
by extracting similarities from nearest neighbor samples to learn enhanced semantic relationships. 
Three data augmentation strategies were introduced to enhance the representation of features extracted by contrastive learning: Band erasure, gradient mask and random occlusion. In the same context, \cite{hu2021deep} introduced 
a spectral-spatial contrastive clustering network, with 
a set of spectral-spatial augmentation techniques that includes random cropping, resizing, rotation, flipping, and blurring for the spatial domain, as well as band permutation and band erasure for the spectral domain. 
Similarly, we proposed in our paper \cite{dhaini2024contrastive} a self-supervised contrastive learning method designed for regression on hyperspectral imagery, featuring custom augmentations (e.g., atmospheric simulation) and a modified contrastive loss employing radius-based pairing. The framework achieved notable performance gains in tasks such as hyperspectral unmixing and pollution concentration prediction \cite{dhaini2021hyperspectral}.

Parallel to these contrastive learning works, there were several studies focusing on spectral-spatial processing through the use of 3D-CNN architectures. A 3D-CNN for classification was introduced in \cite{li2017spectral}, including two 3D convolutional layers. 
Moreover, \cite{luo2018hsi} proposed a 3D architecture composed of one 3D-CNN layer followed by 2D-CNN ones. The 3D-CNN layer is adequate for extracting spectral-spatial features, since it processes the data in three dimensions. Following this, the 2D-CNN layers focus on further extracting and refining spatial features from the spectrally processed data. 
A sophisticated 3D-CNN model was introduced in \cite{hamida20183}. It begins with a 3D convolutional layer employing a 
kernel to capture spatial and spectral features, followed by a pooling layer that reduces the spectral dimension
. The model then replicates this structure with increased neurons for deeper feature extraction. Subsequent layers progressively reduce the spatial dimensionality, enhancing feature abstraction. The architecture proposed in \cite{atik2024dual} leverages two parallel streams to separately process spectral and spatial information. The spectral stream employs 3D convolutions tailored to capture joint spectral–spatial correlations, while the spatial stream refines neighborhood structures. A transformer-based design was introduced in \cite{varahagiri20243d} by combining 3D convolutions for early spectral–spatial embedding with a vision transformer backbone. The initial convolutional blocks extract localized spectral–spatial features, which are then tokenized into patch representations. Multi-head self-attention layers model long-range dependencies across both spectral and spatial dimensions, while projection layers adapt the transformer outputs for different 
tasks, offering improved feature abstraction compared to purely convolutional approaches.

\section{Proposed Framework}
\label{sec:method}

\subsection{Architecture}

Due to the high dimensionality of hyperspectral images, processing the entire hyperspectral cube directly (all wavelengths for all pixels) is computationally intensive and often impractical. To address these challenges, the hyperspectral cube is often divided into smaller, more manageable segments called patches \cite{li2017spectral,wang2020learning}. This division into patches allows for more efficient processing and helps in capturing both spatial and spectral information within a localized region. By focusing on patches, algorithms can extract spatial features (like texture and shape) and spectral features (like reflectance values at different wavelengths) more effectively and use these features to classify each patch or pixel within the patch. Regarding label assignment of the extracted patch, as we are dealing with regression tasks, the average label of the extracted patch can serve as a robust representation of the material in that patch.

\begin{figure}
    \centering
    \includegraphics[width=\columnwidth]{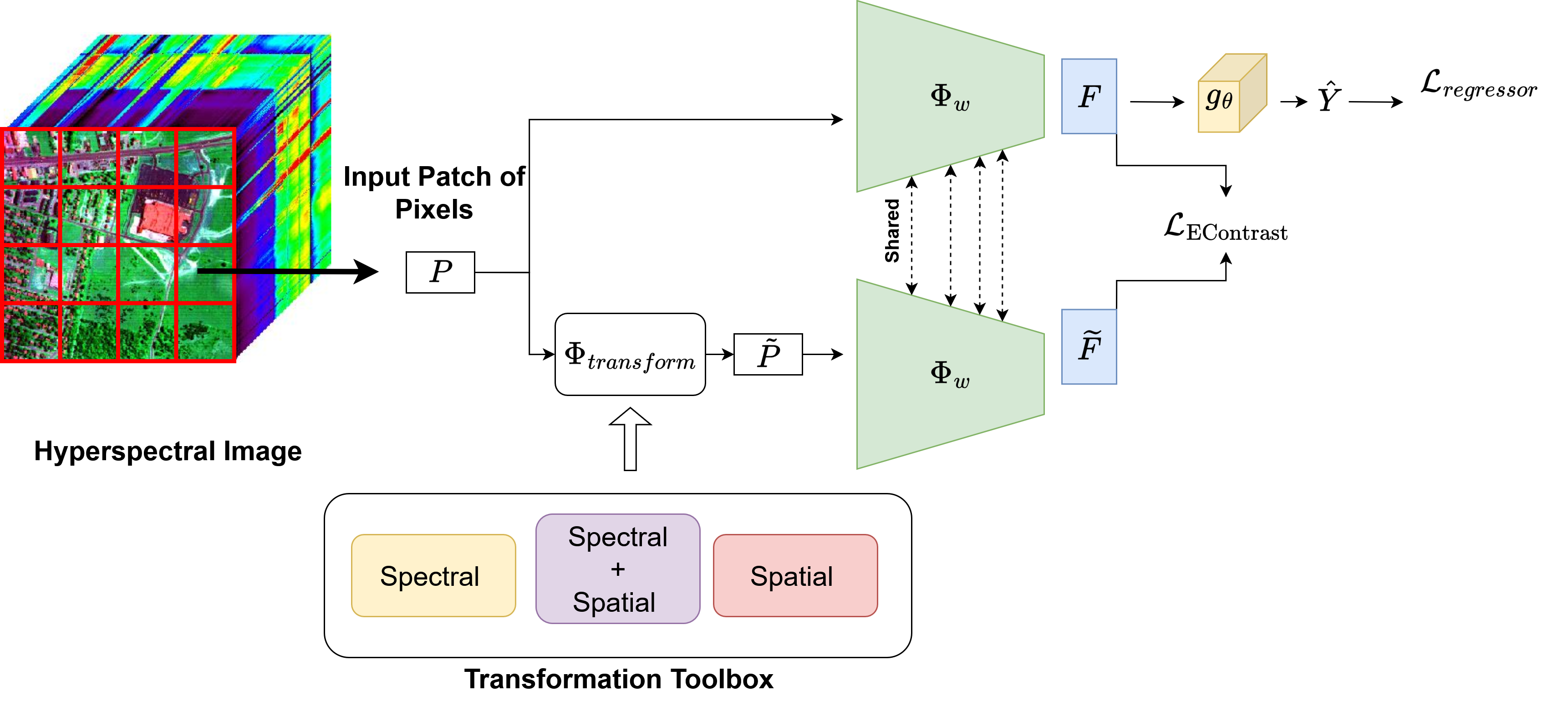}
    \caption{Architecture of the proposed framework.}
    \label{fig:archi}
\end{figure}

 For an input hyperspectral image $\mathbf{X} \in \mathbb{R}^{k \times H \times W}$, with $H$ and $W$ spatial dimensions and  $k$  the total number of wavelengths, a patch $\mathbf{P} \in \mathbb{R}^ {k \times S \times S}$ is extracted, where $S$ is the patch size. Every input patch is then augmented using a transformation toolbox $\Phi_{transform}$ containing spectral, spatial and spectral-spatial transformations. Both the original patch $\mathbf{P}$ and the transformed one, denoted $\widetilde{\mathbf{P}}$, are then passed to a shared feature extractor $\Phi_w$ to get $\mathbf{F} = [f^1,f^2,\ldots,f^{N}]^\top$ and $\widetilde{\mathbf{F}}$, respectively, with $N = S \times S$ in this case. In order to fully capture joint spectral-spatial information, the feature extractor $\Phi_w$ is a 3D neural network backbone, which could be any 3D-CNN architecture, such as in \cite{luo2018hsi,hamida20183,atik2024dual}, or a transformer-based one as in \cite{varahagiri20243d}. These features are passed into a regression network $g_\theta$ to generate the regression labels $\hat{\mathbf{Y}} = [\hat{y}^1,\hat{y}^2,\ldots,\hat{y}^{N}]^\top \in \mathbb{R}^{N \times s}$ for $s$ prediction variables. The network is jointly trained with contrastive and regression losses simultaneously. The architecture of the proposed framework is given in \autoref{fig:archi}.

 In the following, we describe in detail the contrastive learning, before presenting the spectral and spatial data augmentation strategies.

\subsection{Contrastive Learning}

After augmenting the input patch $\mathbf{P}$ with a different view $\widetilde{\mathbf{P}}$, both are forwarded to the feature extractor $\Phi_w$, which extracts the corresponding features, ${\mathbf{F}}$ and ${\widetilde{\mathbf{F}}}$  respectively, and then will be later on passed to a regression head $g_{\theta}$. To optimize the feature extractor, we train the neural network so that patches with different views or close regression labels should share similar features in the latent space, while different spectra with different labels should be far away. This type of training can be achieved using a self-supervised contrastive loss \cite{chen2020simple}. 

In classification, selecting similar data pairs (referred to as positive pairs) can be done in a straightforward way, by taking the transformed version of a sample as well as other samples that belong to the same class, while dealing with the rest as negative ones. In regression, as we do not have class labels, we define for the $i$-th sample a ball $\mathbf{B}^i$ of radius $r$ where positive pair $j$ is selected using the $\ell_2$-norm as following:
\begin{equation}
\left \| y^i-y^j\right \| \leq r.
\end{equation}
This allows us to explore, for regression tasks, any contrastive loss from the literature often proposed for classification tasks.

We consider the contrastive loss used in most recent work based on the cross entropy, namely 
{\small
\begin{equation}
\mathcal{L}_{Contrastive}=-\frac{1}{N} \sum_{i= 1}^{2N} \sum_{j \in \mathbf{B}^i} \log \frac{\exp \left(\operatorname{sim}\left(f^i, f^j\right) / \tau\right)}{\sum_{k \not \in \mathbf{B}^i}  \exp \left(\operatorname{sim}\left(f^i, f^k\right) / \tau\right)}.
\label{hard-contrast}
\end{equation}
}%
where $\operatorname{sim}(\boldsymbol{u}, \boldsymbol{v})=\boldsymbol{u}^T \boldsymbol{v} /(\|\boldsymbol{u}\|\|\boldsymbol{v}\|)$ is the cosine similarity between two vectors, and $\tau$ is a temperature scalar. By minimizing this loss, the similarity between $i$-{th} and $j$-{th} samples is maximized while minimizing the similarity between $i$-{th} and $k$-{th} samples. For training, the contrastive loss is combined with a standard mean squared error regression loss according to the following:
\begin{equation}
     \mathcal{L}_{total}=\mathcal{L}_{\text{R}}+ \alpha \ \mathcal{L}_{\text{Contrastive}},
\end{equation}
\begin{equation}
     \mathcal{L}_{\text{R}}=\frac{1}{N} \sum_{i=1}^{N} \left\| {y}^{i} -
     g_{\theta} \left(f^i\right) \right\|^{2}.
\label{loss_regress}
\end{equation}

\subsection{Spectral Data Augmentation Strategies}

\begin{figure}
    \centering
    \includegraphics[width=\columnwidth]{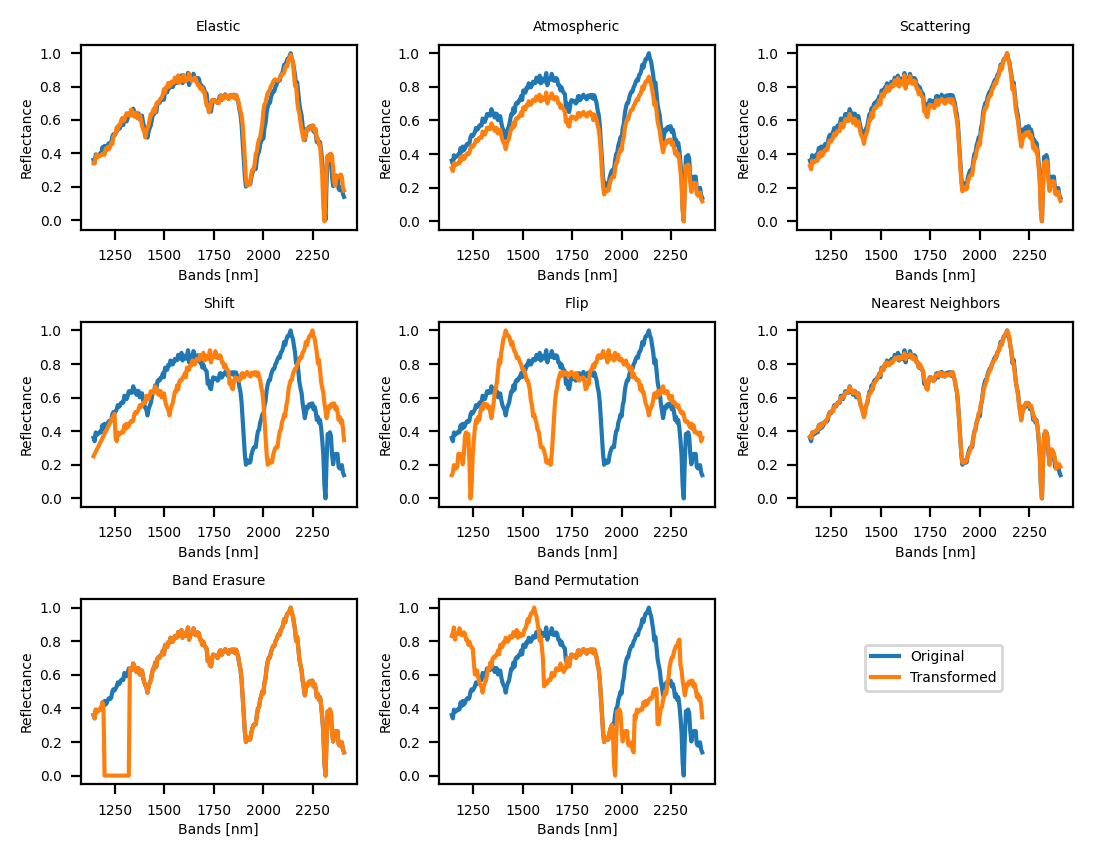}
    \caption{Original (blue) and transformed (orange) examples from real spectral data.}
    \label{fig:tellux-data}
\end{figure}

Standard image augmentation techniques, such as grayscale and color jittering, are not suitable for addressing the spectral domain. Following our previous work \cite{dhaini2024contrastive}, we explore a set of spectral domain augmentation strategies specifically designed for hyperspectral data:
\begin{enumerate}
    \item \textbf{Spectral Shift:} random displacement of the spectrum along the wavelength axis, making the model more robust to wavelength variations.
    \item \textbf{Spectral Flipping:} reversing the order of spectral bands to improve invariance to potential inconsistencies in band ordering across sensors.
    \item \textbf{Scattering Hapke's Model \cite{hapke1981bidirectional}:} simulating realistic scattering effects caused by light--surface interactions.
    \item \textbf{Atmospheric Compensation \cite{uezato2016novel}:} applying radiometric corrections to mimic atmospheric effects under varying illumination conditions.
    \item \textbf{Elastic Distortion:} introducing smooth, random wavelength deformations to model misalignments and spectral variability.
\end{enumerate}
Moreover, we consider the following augmentation strategies:
\begin{enumerate}
    \item \textbf{Band Erasure \cite{hu2021deep}:} masking random wavelengths.
    \item \textbf{Band Permutation \cite{hu2021deep}:} shuffling the order of spectral bands.
    \item \textbf{Nearest Neighbor Mixing \cite{wang2023nearest}:} generating synthetic samples from local averages of neighboring spectra.
\end{enumerate}
 An example of these spectral transformations can be seen in \autoref{fig:tellux-data} when applied to real spectral data.

\vspace{-1em}
\subsection{Spatial Data Augmentation Methods}

In addition to spectral-domain strategies, common spatial augmentations can also be applied to hyperspectral cubes by treating each spectral channel as a separate 2D image. As these transformations are well established in the literature, we briefly recall the most relevant ones here:
\begin{enumerate}
    \item \textbf{Rotation:} random rotations of image patches by a chosen angle.
    \item \textbf{Elastic Deformation:} smooth spatial distortions based on displacement fields.
    \item \textbf{Flipping:} horizontal or vertical mirroring of patches.
    \item \textbf{Translation:} shifting the patch by fixed pixel offsets in the $x$ and/or $y$ directions.
\end{enumerate}

\begin{table}[t]
\centering
\caption{$R^2$ Scores for Synthetic Data.}
\renewcommand{\arraystretch}{1.3} 
\resizebox{\columnwidth}{!}{%
\begin{tabular}{@{}lccccc@{}}

&  {\textbf{Luo 
\cite{luo2018hsi}}} &  {\textbf{Hamida 
\cite{hamida20183}}} &  {\textbf{Li 
\cite{li2017spectral}}} &  {\textbf{Atik 
\cite{atik2024dual}}} &  {\textbf{Varahagiri 
\cite{varahagiri20243d}}} \\
\hline
Baseline  & 0.43 $\pm$ 0.03 & 0.40 $\pm$ 0.03 & 0.56 $\pm$ 0.03 & 0.66 $\pm$ 0.02 & 0.81 $\pm$ 0.02 \\
Spectral contrastive          & 0.55 $\pm$ 0.03 & 0.56 $\pm$ 0.03 & 0.72 $\pm$ 0.03 & 0.74 $\pm$ 0.02 & 0.88 $\pm$ 0.02 \\
Spatial contrastive              & \textcolor{red}{0.58 $\pm$ 0.03} & 0.57 $\pm$ 0.03 & 0.74 $\pm$ 0.05 & 0.75 $\pm$ 0.02 & 0.89 $\pm$ 0.02 \\
Spectral+Spatial & \textcolor{red}{0.58 $\pm$ 0.03} & \textcolor{red}{0.59 $\pm$ 0.03} & \textcolor{red}{0.81 $\pm$ 0.04} & \textcolor{red}{0.78 $\pm$ 0.02} & \textcolor{red}{0.93 $\pm$ 0.02} \\
\hline
\end{tabular}%
}
\label{r2-synth}
\end{table}

\begin{table}[t]
\centering
\caption{Mean Absolute Errors for Synthetic Data.}
\renewcommand{\arraystretch}{1.3} 
\resizebox{\columnwidth}{!}{%
\begin{tabular}{@{}lccccc@{}}
&  {\textbf{Luo 
\cite{luo2018hsi}}} &  {\textbf{Hamida 
\cite{hamida20183}}} &  {\textbf{Li 
\cite{li2017spectral}}} &  {\textbf{Atik 
\cite{atik2024dual}}} &  {\textbf{Varahagiri 
\cite{varahagiri20243d}}} \\
\hline
Baseline     & 0.3122 $\pm$ 0.06 & 0.3249 $\pm$ 0.06 & 0.2865 $\pm$ 0.04 & 0.2555 $\pm$ 0.03 & 0.1930 $\pm$ 0.03 \\
Spectral contrastive             & 0.2912 $\pm$ 0.06 & 0.2859 $\pm$ 0.06 & 0.2271 $\pm$ 0.04 & 0.2255 $\pm$ 0.03 & 0.1730 $\pm$ 0.03 \\
Spatial contrastive              & \textcolor{red}{0.2791 $\pm$ 0.06} & 0.2801 $\pm$ 0.06 & 0.2222 $\pm$ 0.04 & 0.2195 $\pm$ 0.03 & 0.1689 $\pm$ 0.03 \\
Spectral+Spatial & \textcolor{red}{0.2791 $\pm$ 0.06} & \textcolor{red}{0.2675 $\pm$ 0.06} & \textcolor{red}{0.1912 $\pm$ 0.04} & \textcolor{red}{0.2050 $\pm$ 0.03} & \textcolor{red}{0.1471 $\pm$ 0.03} \\
\hline
\end{tabular}%
}
\label{mae-synth}
\end{table}

\section{Expermiments \& Results}
\label{sec:results}

\subsection{Synthetic Spatial Hyperspectral Data}

For the synthetic data, four random endmembers were selected from the 
USGS digital spectral library \cite{swayze1993us}. Each endmember is composed of 224 contiguous bands. A total of $100 \times 100$ pixels were generated with abundances following a Dirichlet distribution. Additive zero-mean Gaussian noise was added to the data with a signal-to-noise ratio of 20 dB. We considered a polynomial post-nonlinear mixing of the endmembers, where the nonlinearity is represented by the element-wise~product of two linear mixtures as following:
\begin{equation}
x=M a+Ma\odot Ma + n,
\label{pnmm}
\end{equation}
with $M$, $a$ and $n$ being the endmember matrix, abundances and noise vector, respectively, and $\odot$ denotes the element-wise~product. In addition, the resulting hyperspectral cube was modulated spatially by multiplying it with a 2D Gaussian pattern, adding a realistic spatial variability to the synthetic data. This approach effectively mimics real-world hyperspectral imagery characteristics \cite{kizel2018spatially}. After that, the hyperspectral cube was divided into $10 \times 10$ patches and each patch was labeled with the average mixing coefficients. 
The model was trained to predict these mixing coefficients.

We examined five benchmark 3D-based backbones described in Section~2 \cite{li2017spectral,luo2018hsi,hamida20183,atik2024dual,varahagiri20243d}. The obtained $R^2$ scores and mean absolute errors are given in \autoref{r2-synth} and \autoref{mae-synth}, respectively. For each model, we considered as baseline the entire model trained using the regression loss only, thus without the contrastive loss. After that, we evaluated the effectiveness of each set of transformations (spectral, spatial and spectral-spatial) on improving this baseline. For the spectral-spatial augmentation strategies, the process starts with a spatial transform for the input patch followed by a spectral one. In the training process, a large batch of 256 was used to help the contrastive learning process as it provides many negative samples per anchor, making the model learn to distinguish representations more effectively. All 3D-based backbones were trained using a stochastic gradient descent optimizer. 

It is clear from these results that contrastive learning improved the baseline scores for all the compared models. We can also see the incremental improvement of the framework, where adding spectral to spatial contrastive learning lead to the best regression metrics for each model. The fusion of both kinds of transformations improves robustness against a wide variety of variabilities that might occur in real scenarios. The spectral-spatial transformer \cite{varahagiri20243d} outperformed the other architectures, which highlights the high performance of transformers on hyperspectral tasks compared to standard 3D-CNN architectures, but with higher computational complexity. Moreover, even such transformers can benefit from contrastive learning to obtain better regression metrics.

\subsection{Samson Dataset}

The Samson dataset obtained by the SAMSON sensor \cite{davis2007spatial} is one of the most widely used hyperspectral datasets for hyperspectral unmixing. The original image contains $952 \times 952$ pixels and 156 bands ranging from $0.401$ to $0.889 ~ \mu \mathrm{m}$. The most used scene is a $95 \times 95$ pixel image cropped from the original image. The dataset consists of three main endmembers:  Soil, Tree, and  Water. The same process as the one presented in the synthetic data section was done. The results are shown in \autoref{r2-samson} and \autoref{mae-samson}. Again we see similar results as obtained in synthetic data, where contrastive learning framework improved the baseline score in every scenario. In addition, the spectral-spatial transformer also obtained the best results, which highlights its performance on real scenes and not only synthetic ones.

\begin{table}[t]
\centering
\caption{$R^2$ Scores for Samson Data.}
\renewcommand{\arraystretch}{1.3} 
\resizebox{\columnwidth}{!}{%
\begin{tabular}{@{}lccccc@{}}
&  {\textbf{Luo 
\cite{luo2018hsi}}} &  {\textbf{Hamida 
\cite{hamida20183}}} &  {\textbf{Li 
\cite{li2017spectral}}} &  {\textbf{Atik 
\cite{atik2024dual}}} &  {\textbf{Varahagiri 
\cite{varahagiri20243d}}} \\
\hline
Baseline 
& 0.59 $\pm$ 0.02 & 0.56 $\pm$ 0.03 & 0.55 $\pm$ 0.03 & 0.65 $\pm$ 0.03 & 0.76 $\pm$ 0.02 \\
Spectral contrastive            & 0.72 $\pm$ 0.02 & 0.68 $\pm$ 0.03 & 0.66 $\pm$ 0.03 & 0.73 $\pm$ 0.03 & 0.86 $\pm$ 0.02 \\
Spatial contrastive             & 0.74 $\pm$ 0.02 & 0.72 $\pm$ 0.03 & 0.69 $\pm$ 0.03 & 0.74 $\pm$ 0.03 & 0.88 $\pm$ 0.02 \\
Spectral+Spatial & \textcolor{red}{0.76 $\pm$ 0.02} & \textcolor{red}{0.75 $\pm$ 0.03} & \textcolor{red}{0.74 $\pm$ 0.03} & \textcolor{red}{0.76 $\pm$ 0.03} & \textcolor{red}{0.91 $\pm$ 0.02} \\
\hline
\end{tabular}%
}
\label{r2-samson}
\end{table}

\begin{table}[t]
\centering
\caption{Mean Absolute Errors for Samson Dataset.}
\renewcommand{\arraystretch}{1.3} 
\resizebox{\columnwidth}{!}{%
\begin{tabular}{@{}lccccc@{}}
&  {\textbf{Luo 
\cite{luo2018hsi}}} &  {\textbf{Hamida 
\cite{hamida20183}}} &  {\textbf{Li 
\cite{li2017spectral}}} &  {\textbf{Atik 
\cite{atik2024dual}}} &  {\textbf{Varahagiri 
\cite{varahagiri20243d}}} \\
\hline
Baseline    & 0.48 $\pm$ 0.009 & 0.50 $\pm$ 0.011 & 0.52 $\pm$ 0.012 & 0.46 $\pm$ 0.012 & 0.34 $\pm$ 0.009 \\
Spectral contrastive             & 0.39 $\pm$ 0.009 & 0.43 $\pm$ 0.011 & 0.45 $\pm$ 0.012 & 0.37 $\pm$ 0.011 & 0.25 $\pm$ 0.009 \\
Spatial contrastive             & 0.36 $\pm$ 0.009 & 0.39 $\pm$ 0.011 & 0.41 $\pm$ 0.011 & 0.36 $\pm$ 0.011 & 0.23 $\pm$ 0.009 \\
Spectral+Spatial & \textcolor{red}{0.34 $\pm$ 0.009} & \textcolor{red}{0.35 $\pm$ 0.011} & \textcolor{red}{0.36 $\pm$ 0.011} & \textcolor{red}{0.34 $\pm$ 0.011} & \textcolor{red}{0.21 $\pm$ 0.009} \\
\hline
\end{tabular}%
}
\label{mae-samson}
\end{table}

\section{Conclusion}
\label{sec:conclusion}

In this paper, we introduced a spectral-spatial contrastive learning framework. 
Moreover, we presented a set of spectral and spatial transformations adequate for hyperspectral data. The relevance of the proposed framework was demonstrated on synthetic and real world datasets. 
Future work will involve combining the presented framework with domain adaptation to generalize knowledge on unseen domains.

\vfill\pagebreak

\bibliographystyle{IEEEbib}
\bibliography{refs}

@article{braham2025spectralearth,
  title={Spectralearth: Training hyperspectral foundation models at scale},
  author={Braham, Nassim Ait Ali and Albrecht, Conrad M and Mairal, Julien and Chanussot, Jocelyn and Wang, Yi and Zhu, Xiao Xiang},
  journal={IEEE Journal of Selected Topics in Applied Earth Observations and Remote Sensing},
  year={2025},
  publisher={IEEE}
}

@article{roy2019hybridsn,
  title={HybridSN: Exploring {3-D--2-D CNN} feature hierarchy for hyperspectral image classification},
  author={Roy, Swalpa Kumar and Krishna, Gopal and Dubey, Shiv Ram and Chaudhuri, Bidyut B},
  journal={IEEE Geoscience and Remote Sensing Letters},
  volume={17},
  number={2},
  pages={277--281},
  year={2019},
  publisher={IEEE}
}

@article{dhaini2022end,
  title={End-to-End Convolutional Autoencoder for Nonlinear Hyperspectral Unmixing},
  author={Dhaini, Mohamad and Berar, Maxime and Honeine, Paul and Van Exem, Antonin},
  journal={Remote Sensing},
  volume={14},
  number={14},
  pages={3341},
  year={2022},
  publisher={MDPI}
}

@inproceedings{chen2020simple,
  title={A simple framework for contrastive learning of visual representations},
  author={Chen, Ting and Kornblith, Simon and Norouzi, Mohammad and Hinton, Geoffrey},
  booktitle={International conference on machine learning},
  pages={1597--1607},
  year={2020},
}

@article{wang2023nearest,
  title={Nearest Neighbor-Based Contrastive Learning for Hyperspectral and {LiDAR} Data Classification},
  author={Wang, Meng and Gao, Feng and Dong, Junyu and Li, Heng-Chao and Du, Qian},
  journal={IEEE Transactions on Geoscience and Remote Sensing},
  year={2023},
  publisher={IEEE}
}

@article{hu2021deep,
  title={Deep spatial-spectral subspace clustering for hyperspectral images based on contrastive learning},
  author={Hu, Xiang and Li, Teng and Zhou, Tong and Peng, Yuanxi},
  journal={Remote Sensing},
  volume={13},
  number={21},
  pages={4418},
  year={2021},
  publisher={MDPI}
}

@article{hapke1981bidirectional,
  title={Bidirectional reflectance spectroscopy: 1. Theory},
  author={Hapke, Bruce},
  journal={Journal of Geophysical Research: Solid Earth},
  volume={86},
  number={B4},
  pages={3039--3054},
  year={1981},
  publisher={Wiley Online Library}
}

@article{uezato2016novel,
  title={A novel endmember bundle extraction and clustering approach for capturing spectral variability within endmember classes},
  author={Uezato, Tatsumi and Murphy, Richard J and Melkumyan, Arman and Chlingaryan, Anna},
  journal={IEEE Transactions on Geoscience and Remote Sensing},
  volume={54},
  number={11},
  pages={6712--6731},
  year={2016},
  publisher={IEEE}
}

@inproceedings{swayze1993us,
  title={The US Geological Survey, Digital Spectral Library: Version 1: 0.2 to 3.0 mum},
  author={Swayze, GA and Clark, RN and King, TVV and Gallagher, A and Calvin, WM},
  booktitle={Bulletin of the American astronomical society},
  volume={25},
  pages={1033},
  year={1993}
}

@inproceedings{dhaini2021hyperspectral,
  title={Hyperspectral imaging for the evaluation of lithology and the monitoring of hydrocarbons in environmental samples},
  author={Dhaini, Mohamad and Roudaut, Fran{\c{c}}ois-Joseph and Garret, Antonin and Arzur, Ronan and Chereau, Audrey and Buhler-Varenne, Fanny and Honeine, Paul and Mignot, M{\'e}lanie and van Exem, Antonin},
  booktitle={RemTech (International event on Remediation, Coasts, Floods, Climate, Seismic, Regeneration Industry)},
  year={2021}
}

@article{li2017spectral,
  title={Spectral--spatial classification of hyperspectral imagery with {3D} convolutional neural network},
  author={Li, Ying and Zhang, Haokui and Shen, Qiang},
  journal={Remote Sensing},
  volume={9},
  number={1},
  pages={67},
  year={2017},
  publisher={MDPI}
}

@inproceedings{luo2018hsi,
  title={{HSI-CNN}: A novel convolution neural network for hyperspectral image},
  author={Luo, Yanan and Zou, Jie and Yao, Chengfei and Zhao, Xiaosong and Li, Tao and Bai, Gang},
  booktitle={International Conference on Audio, Language and Image Processing (ICALIP)},
  pages={464--469},
  year={2018},
  organization={IEEE}
}

@article{hamida20183,
  title={{3-D} deep learning approach for remote sensing image classification},
  author={Hamida, Amina Ben and Benoit, Alexandre and Lambert, Patrick and Amar, Chokri Ben},
  journal={IEEE Transactions on geoscience and remote sensing},
  volume={56},
  number={8},
  pages={4420--4434},
  year={2018},
  publisher={IEEE}
}

@article{atik2024dual,
  title={Dual-stream spectral-spatial convolutional neural network for hyperspectral image classification and optimal band selection},
  author={Atik, Saziye Ozge},
  journal={Advances in Space Research},
  volume={74},
  number={5},
  pages={2025--2041},
  year={2024},
  publisher={Elsevier}
}

@inproceedings{varahagiri20243d,
  title={{3D}-convolution guided spectral-spatial transformer for hyperspectral image classification},
  author={Varahagiri, Shyam and Sinha, Aryaman and Dubey, Shiv Ram and Singh, Satish Kumar},
  booktitle={2024 IEEE Conference on Artificial Intelligence (CAI)},
  pages={8--14},
  year={2024},
  organization={IEEE}
}

@inproceedings{dhaini2024contrastive,
  title={Contrastive learning for regression on hyperspectral data},
  author={Dhaini, Mohamad and Berar, Maxime and Honeine, Paul and Van Exem, Antonin},
  booktitle={IEEE International Conference on Acoustics, Speech and Signal Processing (ICASSP)},
  pages={5080--5084},
  year={2024},
  organization={IEEE}
}

@article{shenming2022new,
  title={A new hyperspectral image classification method based on spatial-spectral features},
  author={Shenming, Qu and Xiang, Li and Zhihua, Gan},
  journal={Scientific Reports},
  volume={12},
  number={1},
  pages={1541},
  year={2022},
  publisher={Nature Publishing Group UK London}
}

@inproceedings{wang2020learning,
  title={Learning spectral-spatial prior via {3DDNCNN} for hyperspectral image deconvolution},
  author={Wang, Xiuheng and Chen, Jie and Richard, C{\'e}dric and Brie, David},
  booktitle={IEEE International Conference on Acoustics, Speech and Signal Processing (ICASSP)},
  pages={2403--2407},
  year={2020},
  organization={IEEE}
}

@article{kizel2018spatially,
  title={Spatially adaptive hyperspectral unmixing through endmembers analytical localization based on sums of anisotropic {2D} Gaussians},
  author={Kizel, Fadi and Shoshany, Maxim},
  journal={ISPRS Journal of Photogrammetry and Remote Sensing},
  volume={141},
  pages={185--207},
  year={2018},
  publisher={Elsevier}
}

@inproceedings{davis2007spatial,
  title={Spatial and spectral resolution considerations for imaging coastal waters},
  author={Davis, Curtiss O and Kavanaugh, Maria and Letelier, Ricardo and Bissett, W Paul and Kohler, David},
  booktitle={Coastal Ocean Remote Sensing},
  volume={6680},
  pages={196--207},
  year={2007},
  organization={SPIE}
}

\end{document}